# Exoskeleton-Based Multimodal Action and Movement Recognition: Identifying and Developing the Optimal Boosted Learning Approach

**Nirmalya Thakur** THAKURNA@MAIL.UC.EDU

*Department of Electrical Engineering and Computer Science,*
*University of Cincinnati, Cincinnati, OH 45221-0030, USA.*

**Chia Y. Han** HAN@UCMAIL.UC.EDU

*Department of Electrical Engineering and Computer Science,*
*University of Cincinnati, Cincinnati, OH 45221-0030, USA.*

**Corresponding Author:** Nirmalya Thakur.

## Abstract

This paper makes two scientific contributions to the field of exoskeleton-based action and movement recognition. First, it presents a novel machine learning and pattern recognition-based framework that can detect a wide range of actions and movements - walking, walking upstairs, walking downstairs, sitting, standing, lying, stand to sit, sit to stand, sit to lie, lie to sit, stand to lie, and lie to stand, with an overall accuracy of 82.63%. Second, it presents a comprehensive comparative study of different learning approaches - Random Forest, Artificial Neural Network, Decision Tree, Multiway Decision Tree, Support Vector Machine, k-NN, Gradient Boosted Trees, Decision Stump, AutoMLP, Linear Regression, Vector Linear Regression, Random Tree, Naïve Bayes, Naïve Bayes (Kernel), Linear Discriminant Analysis, Quadratic Discriminant Analysis, and Deep Learning applied to this framework. The performance of each of these learning approaches was boosted by using the AdaBoost algorithm, and the Cross Validation approach was used for training and testing. The results show that in boosted form, the k-NN classifier outperforms all the other boosted learning approaches and is, therefore, the optimal learning method for this purpose. The results presented and discussed uphold the importance of this work to contribute towards augmenting the abilities of exoskeleton-based assisted and independent living of the elderly in the future of Internet of Things-based living environments, such as Smart Homes. As a specific use case, we also discuss how the findings of our work are relevant for augmenting the capabilities of the Hybrid Assistive Limb exoskeleton, a highly functional lower limb exoskeleton.

## 1. INTRODUCTION

In 1950, 8% of the world's total population was considered the elderly population and in 2000,the population count of the elderly increased to 10% of the total population. Their population is expected to increase to 21% by 2050 [1]. This demonstrates the unprecedented rate at which the elderly population of the world is increasing. The process of aging is associated with multiple needs on a day-to-day basis, specifically for performing Activities of Daily Living (ADLs). There exist multiple definitions of ADLs. A broad definition of ADL may be obtained from the work in [2], which defines ADL as the tasks and activities that a person performs for sustaining themselves on a regular and routine basis in their living environments. ADLs may be grouped into five broad types of activities: Personal Hygiene, Dressing, Eating, Maintaining Continence, and Mobility. ADLs can occur in multiple ways based on the associated spatial and temporal features of the environment as well as the user's need, motive, intention, desire, and goals [2]. The ability to walk and move independently decreases with age due to decreased motor neuron, muscle fiber, and aerobic capacity, which limits the independence of elderly people in performing these activities [3]. These limitations in the abilities related to walking and moving also cause a decrease in stride length and muscular strength in older adults. A typical example is older adults tend to walk slowly as compared to their younger counterparts, and falls are also common in the elderly on account of the above reasons [4]. Crutches, manual wheelchairs, power wheelchairs, and several advancements related to the same have been proposed as solutions over the years to help with the independence of the elderly for moving around and for performing ADLs, but there exist several limitations that come in the way of the widescale use and adaptation of such technologies and tools [5, 6].

In consideration of the above challenges, several researchers in the recent past have investigated exoskeletons as a navigation tool for older adults to help them perform their routine-based tasks and activities in the future of technology-driven living spaces; one common example of the same is Smart Homes.

In a broad manner, an exoskeleton may be defined as a *wearable* solution that increases the wearer's performance in one or more ways with a specific focus on improving their abilities such as speed, strength, and endurance [7]. Whenever the wearer has to perform any movement or activity, they exert a force on the exoskeleton. The exoskeleton system detects this force and the associated control signals. Thereafter, it amplifies the same so that the wearer can use the required force for performing the desired movement or activity [7]. Exoskeletons have a wide range of potential applications [8] for not only the independent and assisted living of the elderly but also for those with multiple types of disabilities. Exoskeleton-based solutions are also becoming popular in various application domains: in the military – for decreasing fatigue and increasing productivity; in healthcare – for improving the quality of life of individuals who have lost one or both their legs or arms; in firefighting – to enable firefighters in climbing as well as for lifting of heavy equipment; and in various physical labor-intensive industries – for increasing worker productivity and for reducing the chances of injury during labor-intensive tasks [8]. Because of this range of potential applications, the exoskeleton market worth was about $200 million in 2017 [9] and is estimated to reach $1.3 billion by 2024 [10].

For the future of exoskeletons to contribute towards the assisted living of the elderly with a specific focus on their independence during ADLs, it is necessary that such solutions can detect the differences between different kinds of actions and movements associated with ADLs to assist the user in an action-centered and movement-centered manner. Such intelligent detection and adaptation of exoskeletons in terms of increasing the wearer's abilities as per the specific action or movement being carried out would make these exoskeleton-based solutions more reliable and usable as well as it would help to contribute towards their increased acceptance amongst the elderly. The main motivation behind the proposed work discussed in this paper was to address this research challenge. The review of state of the art and emerging works in this field is discussed in section 2. Section 3 outlines the exoskeleton technology and provides details about a potential real-world application of our proposed framework by discussing how it can be integrated into the Hybrid Assistive Limb exoskeleton, a commercially available lower limb exoskeleton. In Section 4, we discuss the methodology for the development of the proposed intelligent framework that can detect multiple movements and behavioral patterns associated with ADLs. This includes but is not limited to walking, walking up and down the stairs, sitting, standing, lying, and compound movements such as stand to sit, sit to stand, sit to lie, lie to sit, stand to lie, and lie to stand. In this section, the methodology for the comparative study that included developing and testing 17 different machine learning models to determine the ideal learning model for this application is also outlined. Section 5 presents the novel findings of this research work. In section 6, we summarize the paper by outlining the scientific contributions of this work and outline the scope for future work in this field.

## 2. LITERATURE REVIEW

Currently, wearable devices such as exoskeleton-based solutions are still in their infancy. More affordable and available assistive devices consist of canes, axially or underarm crutches used for a temporary recovery period, or wheelchairs for long-term solutions. A multitude of assistive devices and applications, some as common as technology-based crutches and wheelchairs, have been scrutinized throughout the last few years to support the independent living of the elderly with a strong focus on ADLs. Engel et al. [12] developed a smart cane that could monitor Partial Weight Bearing (PWB) signals by the use of a compressible spring within the tube of the cane. The force of the magnitude applied by the user to the spring triggered micro switches in the system. An intelligent sensing mechanism was incorporated in a crutch by Bergmann et al. [13]. This mechanism could detect the force exerted by the user on a forearm crutch to infer the associated gait and its characteristics. An intelligent cane was developed by Wu et al. [14], specifically for the elderly, that used sensor technologies to track the user's well-being. The data collected by the sensors were wirelessly communicated to a PDA by Bluetooth technology for interpretation and analysis. Šantić et al. [15] proposed a crutch-based solution that could monitor the forces exerted by the user on the system by use of an infrared transducer and from force sensors strapped to the user's shoes. This data was used by the system to infer the dynamics of the user's gait and requirements. In a similar manner, other researchers have also focused on tracking and interpreting the weight exerted by a user on their shoes to infer the dynamics of their gait and movement characteristics [16, 17].

In [18], the authors proposed a framework that could track the joint angles, angular velocity, and force/torque by the use of sensor technologies to differentiate between standing, walking, and sitting so that exoskeleton-based solutions can help the user as per the specific motion being carried out by them. In [19], the researchers evaluated the user's joint trajectories in order to detect various categories of motions, and in [20], to infer gait characteristics, the researchers studied the dynamics of motions linked with the user's leg during different activities. Lim et al. [21] created a neural network-based method to infer gait dynamics using multiple data factors such as age, gender, height, and weight to train the model. The model's output provided details on walking speed and frequency by taking into consideration each of these factors. Jonic et al. [22] used machine learning approaches to predict muscle activation patterns with the use of sensor technologies for an exoskeleton to assist the wearer in the walking motion.

Madarasz et al. [23] proposed an intelligent wheelchair-based solution that comprised of a micro-computer, a camera, and an ultrasound scanner. These three units of the system worked together to allow safe navigation of the user in populated environments without the need for any assistance from a caregiver. Miller et al. [24] developed a wheelchair-based system called Tin Man I that consisted of three functionalities to assist in the independent movement of the user. These functionalities included obstacle deviation and the ability to move the wheelchair to a specific point by providing X and Y-coordinate information. As an upgrade to this system, in [25], Miller et al. incorporated functionalities such as storing travel information, returning to the starting point, following walls, passing through doors, and recharging the battery. A smart wheelchair with two extra legs was proposed by Wellman et al. [26]. The chair's additional legs made it possible for it to climb stairs and traverse difficult terrain. In addition to these features, researchers have studied wheelchair-based remedies for quadriplegics, in which facial expression recognition was utilized to direct the wheelchair to a specified place [27-29].

Despite the above advances in this field, several challenges remain [5, 6]. These include the following:

(1) Crutches have several limitations. These include - (a) Crutches limit upper-body freedom. (b) Strong body balance is necessary for patients who use axillary or so-called underarm crutches. (c) Crutches can cause strain and injuries on the arms and upper body, which could lead to paralysis in some cases. (d) Improper pressure on nerves in the arms due to using crutches in the wrong manner can cause loss of balance. (e) Older adults with low upper body strength find it difficult to use crutches.

(2) Manual wheelchairs have multiple limitations. These include – (a) The person in the wheelchair needs assistance for movement. (b) Body strength is required for moving the person in the wheelchair. (c) They can cause minor to major strains, specifically in the shoulders of the person using the wheelchair. (d) Difficult to use in paths with an upward incline as well as for long-distance movements.

(3) Smart and powered wheelchairs have several disadvantages. These include – (a) They are costly. (b) They weigh more as compared to manual wheelchairs. (c) Power usage is a concern. (d) Frequent maintenance is necessary. (e) Familiarization with the controls can be daunting to some elderly.

(4) The existing solutions do not consist of the methodology to detect the comprehensive range of actions and movements associated with ADLs, so action-specific and movement-specific adaptation of an exoskeleton is difficult to achieve.

(5) The most favorable learning method that needs to be implemented by specific behavior application-based exoskeletons is still not known. This is due to the limitation in prior works, which did not compare various machine learning methods to determine the ideal learning method for the detection of actions and movements related to ADLs.

(6) Detection of such actions and movements should be performed by the future of exoskeletons with high levels of accuracy. The existing approaches and frameworks did not investigate any approaches for improving or boosting the detection of movements and actions during ADLs.

We aim to address the above challenges in this field in our paper. The work presented in our paper takes an interdisciplinary approach by integrating the state-of-the-art technologies and advances from different disciplines of Computer Science.

## 3. DISCUSSION OF THE EXOSKELETON TECHNOLOGY AND OVERVIEW OF THE HYBRID ASSISTIVE LIMB EXOSKELETON

In this section, we briefly discuss the exoskeleton technology with a specific focus on lower limb exoskeletons. In a broad manner, our framework, in general, can be beneficial in increasing functionalities of different kinds of exoskeleton-based assistive solutions, but it is expected to be most relevant for enhancing the effectiveness of the Hybrid Assistive Limb (HAL) exoskeleton [11],
a type of commercially available lower-limb exoskeleton that is highly functional and supports multiple applications as well as a range of use cases. Therefore, we also discuss the working of the HAL exoskeleton in detail in this section as well as outline how our framework can support to augment its abilities.

Exoskeleton technology comes in many forms in terms of the design, functionality, operation, and the specific application for which it is being used. An exoskeleton's design is key to its functionality. The design process is very complicated and demands accuracy and precision owing to numerous aspects, including energy supply, user interface, sensing capabilities, monitoring, degree of freedom, systems and sub-systems' behavior, and the biomechanics of the system as a whole [30]. As per the findings of [31], the critical factors for the configuration of an exoskeleton include an actuator – which is utilized to initiate high torque whenever the exoskeleton uses high velocities; a human-exoskeleton interface – which requires tracking and analysis of various aspects of user interactions to guarantee safe interaction with the exoskeleton system; and design optimization – which includes (1) ergonomic and user-comfort considerations for the exoskeleton system. Exoskeleton systems can be classified according to the specific body part for which they are built - for example, upper limb, lower limb, and specific joint exoskeletons such as the knee, shoulder, elbow, ankle, and so on. [32-34].

Upper limb exoskeletons: Robotics-based assistive devices for upper limbs may be broadly classified into two major categories – exoskeletons and prostheses. Such exoskeletons are used by individuals with disabilities in the upper limb region as well as they help to augment the strength, functionalities, and endurance of the upper body parts of individuals with declining physical abilities, such as the elderly. People utilize prostheses for replacing a lost body part [35]. Examples include Trackhold and AmrmeoSpring, two exoskeletons for the upper limbs which have applications related to dynamic tracking and gravitational correction [36,37].

Lower limb exoskeletons: Such exoskeletons are a mechatronic system that aids in walking, standing, and performing similar movements. These systems can also be broadly divided into two categories – gait training systems - which help to improve and recover walking patterns, and those systems, which are designed for assisting with different kinds of movements other than walking. Recent examples of lower limb exoskeletons include EKSO, developed by Ekso Bionics [38], and LOKOMAT, developed by Hocoma [39].

Research has shown that lower limb exoskeletons offer major possibilities to support and assist various kinds of movements. Usually, the specific structure of a lower leg exoskeleton contains three sections or components - the thigh section, the shank section, and the foot section. Specialized joints link these sections, which can detect and quantify the motions of the user's biological joints [40-42]. To encompass the range of motion for the knee and ankle about both the frontal and sagittal planes, a degree of freedom is often added. This additional degree of freedom allows unrestricted movements in the frontal plane without interfering with the measurement of the joint sensors [42]. This range of features and capabilities of lower limb exoskeletons serves as our motivation for discussing how our framework can be integrated into the same. For a specific use case of the framework presented in this paper, we outline how the same can be applied to expanding the functionalities of the Hybrid Assistive Limb (HAL) [11]. HAL is a commercially available lower-limb exoskeleton that was initially developed to assist people in performing their ADLs and labor-intensive tasks. HAL is available as a single leg version which the user needs to wear around their waist and on their affected leg by using the built-in straps fixed on the thigh and shank segments to attach the exoskeleton to their body. HAL-1, the earliest version of HAL, could amplify the wearers walking ability by amplifying the wearer's joint torque. An upgrade on the same was HAL-3, which takes a multimodal approach to assist the wearer in performing ADLs. However, no work thus far has focused on augmenting the abilities of HAL by enabling it to detect the multimodal actions and movements associated with ADLs so that the exoskeleton can take a personalized, user-centered, activity and movement-specific approach towards assisting the wearer during ADLs. Our proposed framework holds the potential to address this research gap – which is further discussed in this section and in the upcoming sections.

The HAL exoskeleton works by using a Cybernic Voluntary Control (CVC) and a Cybernic Autonomous Control (CAC) that enables the exoskeleton system to support wearers based on their specific needs associated with movements by combining the outcome and performance of two algorithms that are specifically meant for these controls [11]. The CVC provides physical support based on the wearer's intention, which is detected by bioelectrical signal-based data, specifically the myoelectricity data, that the system obtains from muscle activity. The CVC can track this data from multiple joints over the entire lower limb; therefore, HAL is able to support healthy individuals in need of additional strength for specific activities, elderly people with declining motor strength, as well as disabled individuals with limited abilities in their lower limbs. The CAC generates a functional motion as per the wearer's body constitution, conditions, and purposes of motion support. It uses joint reaction force and joint angle data to provide comfortable physical support to the wearer. The HAL uses a combination of active and passive joints for its operation. The ankle receives passive power while the knee and hip joints receive active power from a DC motor with a harmonic drive that is located at these joints. The system architecture of HAL consists of multiple sensors that include skin-surface electromyographic (EMG) electrodes placed below the hip and above the knee on both the front and the back sides of the wearer's body, potentiometers for joint angle measurement, ground reaction force sensors, and a gyroscope and accelerometer mounted on the backpack for torso posture estimation. Out of all these sensors, the accelerometer sensors are most relevant to us as our framework used different characteristics of accelerometer data, and therefore it can be developed and integrated into the HAL to expand its functionalities.

In the following section, the system architecture and steps associated with the development of this framework are presented and discussed.

## 4. METHODOLOGY

Any personalized assistive device to elderly, as we envision for the future, will consist of some form of 'smart' exoskeleton, which is adaptable to the physical dimensions of the human subject and able to enhance the body mobility based on the parametric data corresponding to the wearer's bodily movements in the context of small tasks required for conducting basic and necessary daily activities. We propose this data-centric intelligent framework to allow both the capture of body motion data from wearable sensors and to use these data in the activation of assistive devices. In a broad manner, our framework, in general, can be beneficial in increasing functionalities of different kinds of exoskeleton-based assistive solutions, but it is expected to be most relevant for enhancing the effectiveness of the Hybrid Assistive Limb (HAL) exoskeleton [11], which was outlined in Section 3.

This section presents the methodology and the system architecture for the development of our machine-learning and pattern-recognition-based framework that can detect various behavioral patterns related to ADLs. We also present the details about the methodology that we followed for performing the comparative study, where we compared 17 machine learning approaches to determine the most efficient machine learning method for the proposed application.

This system architecture of the proposed framework has a structure that distinguishes motion and activity by detecting, analyzing, and assessing the accelerometer data. We were keen to see the data from the accelerometer as accelerometers play a vital role in the HAL exoskeleton's design [11]. In our lab, we have setup and deployed a real-time system to collect IoT-based Big Data related to human behavior and user activities with various environmental contexts for several ADLs related to microwave usage, reading, working, resting, and learning. This data collection system uses wireless sensors and wearables and simulates a technology-based living environment of the future, a typical example of which could be a smart home, for data collection and analysis. In addition, we have received IRB permission from our university to conduct real-time studies with human subjects in our lab in a way that prioritizes the safety of the human subjects involved. However, we couldn't obtain real-time data using this data collection system because of the COVID-19 situation and the accompanying "work from home" recommendations from several government sectors in the United States [44] at the time this research was conducted.

Therefore, to validate and assess our framework and its capabilities, we utilized the dataset created by Reyes-Ortiz et al. [45]. We selected this dataset because its attributes matched the data that we would have otherwise gathered using our data collection system. To get the data and its characteristics, the researchers [45] had attached an android phone to the waist of the participants who participated in the trials and tracked the behavioral-based data from the android phone's built-in accelerometer sensor. The authors recorded the behavioral-based data associated with various actions that were carried out in the experimental trials, which comprised everything from walking, to going up and down the stairs, to sitting, standing, lying, and standing up from sitting, sitting down from standing, lying down from standing, and standing up from lying. A total of 30 participants, consisting of both male and female participants, aged 19-48, participated in these trials. During the experiments, video recording was performed to label the actual movements in the dataset, which consists of a total of 7767 rows, with each row consisting of the characteristics of the user's motion obtained from the wearables.

We performed the data preprocessing steps and thereby developed the machine learning model by using RapidMiner [46]. RapidMiner is a data science platform that allows the development and implementation of various data science, machine learning, and artificial intelligence-related algorithms and applications [46]. The applications developed in RapidMiner can also communicate with other software applications and tools as per the need – this is one of the key advantages of this data science platform. In this study, we used the RapidMiner Studio, version 9.9.000, on a Microsoft Windows 10 machine with an Intel® Core(TM) i7-7600U processor. RapidMiner Studio's free edition has a data processing threshold of 10,000 rows. Therefore, we had to use the RapidMiner Education License. There is no restriction to the number of rows in the dataset that may be processed while using the Education License of RapidMiner. In RapidMiner, two specific terminologies are important – 'process' and 'operator', for any data science, machine learning, or artificial intelligence-related algorithm or application that is developed. An 'operator' represents a specific functionality of an application such as data processing, data filtering, etc. There are several 'operators' that are already available in the RapidMiner Studio and can be directly used for the development of an application with minor to major modifications or customizations as per the need. RapidMiner also allows the development of new 'operators' with one or more functionalities. A combination of 'operators' that represents an application or algorithm is known as a 'process' in RapidMiner.

During the data analysis steps, attributes from the data not considered relevant for the development of the proposed functionalities in the framework were filtered out using a data filter. The attributes that we used for development of this learning model consisted of tBodyAcc-Mean-1, tBodyAcc-Mean-2, tBodyAcc-Mean-3, tBodyAcc-STD-1, tBodyAcc-STD-2, tBodyAcc-STD-3, tBodyAcc-Mad-1, tBodyAcc-Mad-2, tBodyAcc-Mad-3, tBodyAcc-Max-1, tBodyAcc-Max-2, tBodyAcc-Max-3, tBodyAcc-Min-1, tBodyAcc-Min-2, tBodyAcc-Min-3, and activity. For each of these attributes, 'Mean', 'STD', 'Mad', Max', and 'Min' represented the mean value, standard deviation, median absolute value, maximum value, and minimum value, respectively. To add, 'tBodyAcc' meant the acceleration of the total body recorded during each of these movements and activities. The 'activity' column consisted of the different activities as mentioned above.

After performing these data preprocessing steps by using the 'Data Preprocess' operator, which we had developed, we used the built-in 'Set Role' operator to instruct the machine learning model of the attribute that it should be predicting. The relationships of these various attributes, along with the specific activities, were then studied in RapidMiner. FIGURES 1-5 show the relationship between activities and mean, standard deviation, median, maximum, and minimum values of the accelerometer data, respectively.

Thereafter we developed the learning model. We used the k-Nearest Neighbors (k-NN) classifier with the Cross-Validation approach along with the AdaBoost algorithm. The k-NN approach [47] is a machine learning technique that is widely used for regression and classification problems. The algorithm classifies new data points based on similarity characteristics in terms of a distance function, and the classification is done by use of a majority vote from neighbors. Cross-validation is performed in a machine learning model to reduce the influence of random initialization and to get more stable results as well as for evaluating the generalization error of the algorithm, so the mean and standard deviation of all the runs are taken into account to evaluate the behavior of the algorithm [48]. Therefore, the use of cross-validation is recommended for any machine learning model. AdaBoost can be used to boost the performance of any machine learning model. As we are interested in building more accurate solutions – so, we used AdaBoost with the combination of k-NN and cross-validation in our proposed framework. The cross-validation operator used 10 folds, and the number of iterations of AdaBoost was also defined as 10.

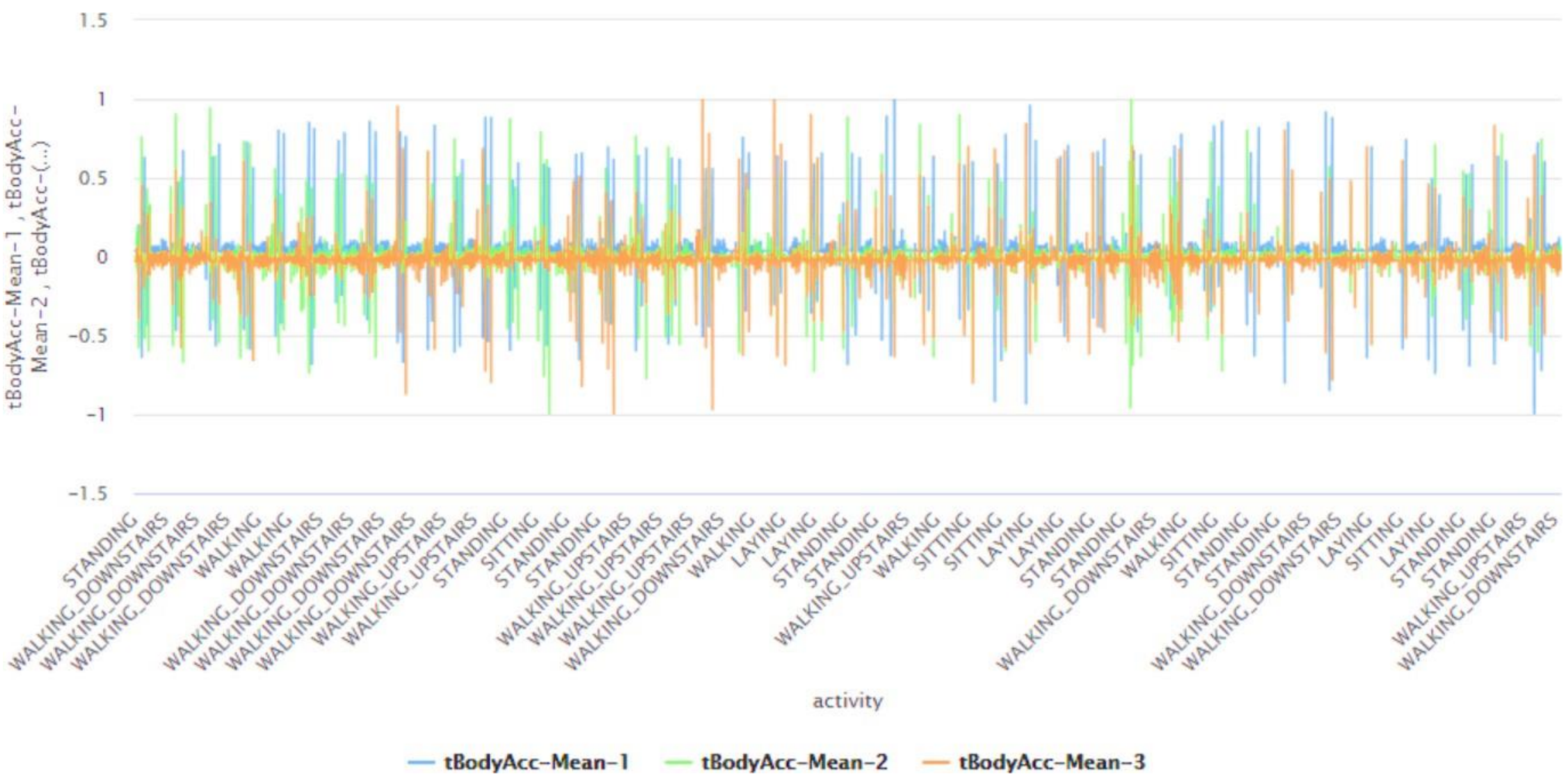

Figure 1: Relationship between tBodyAcc-Mean-1, tBodyAcc-Mean-2, and tBodyAcc-Mean-3 with the different activities. For clarity of representation, only a few activity instances from the dataset are shown here.

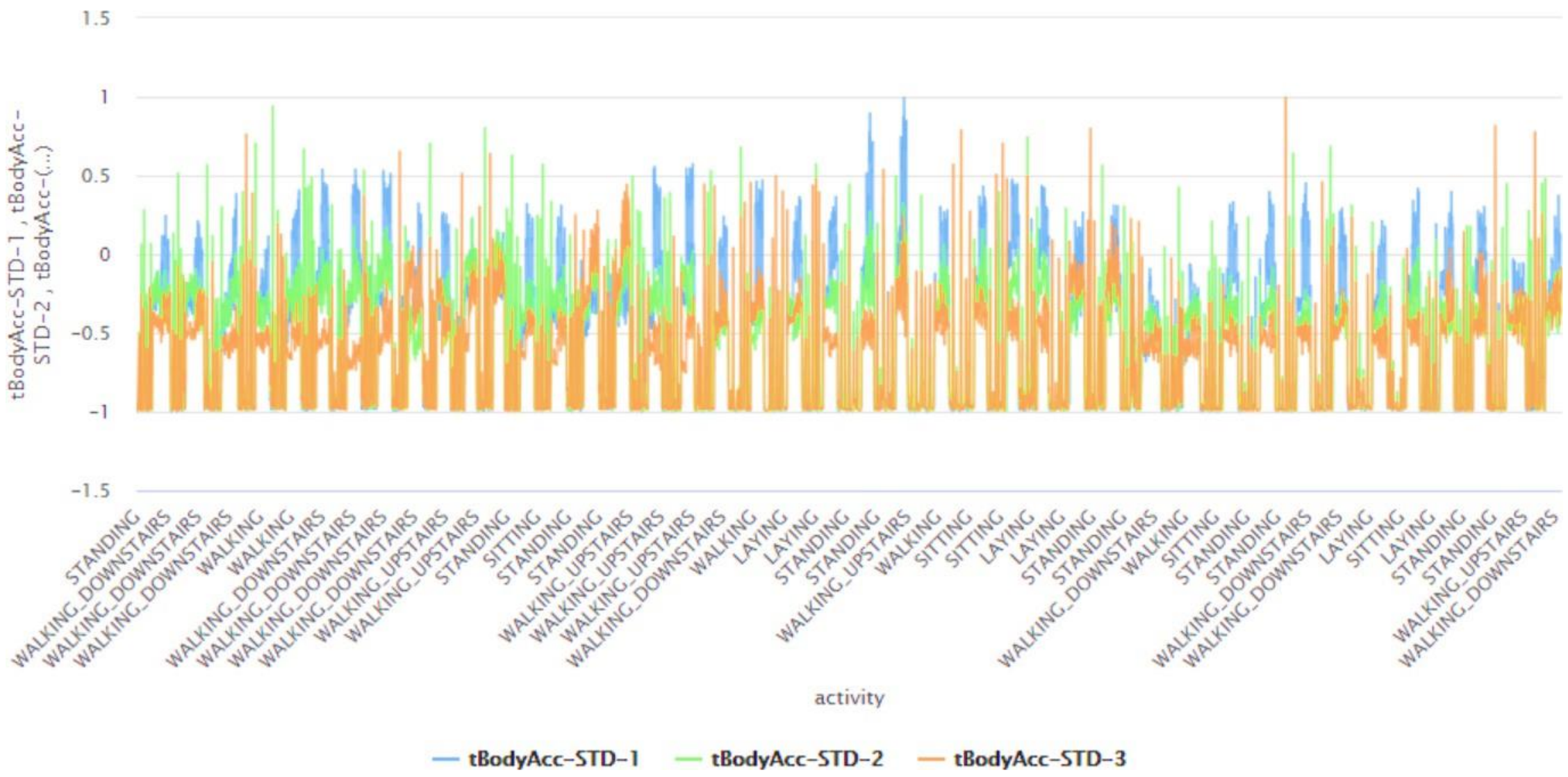

Figure 2: Relationship between tBodyAcc-STD-1, tBodyAcc-STD-2, tBodyAcc-STD-3 with the different activities. For clarity of representation, only a few activity instances from the dataset are shown here.

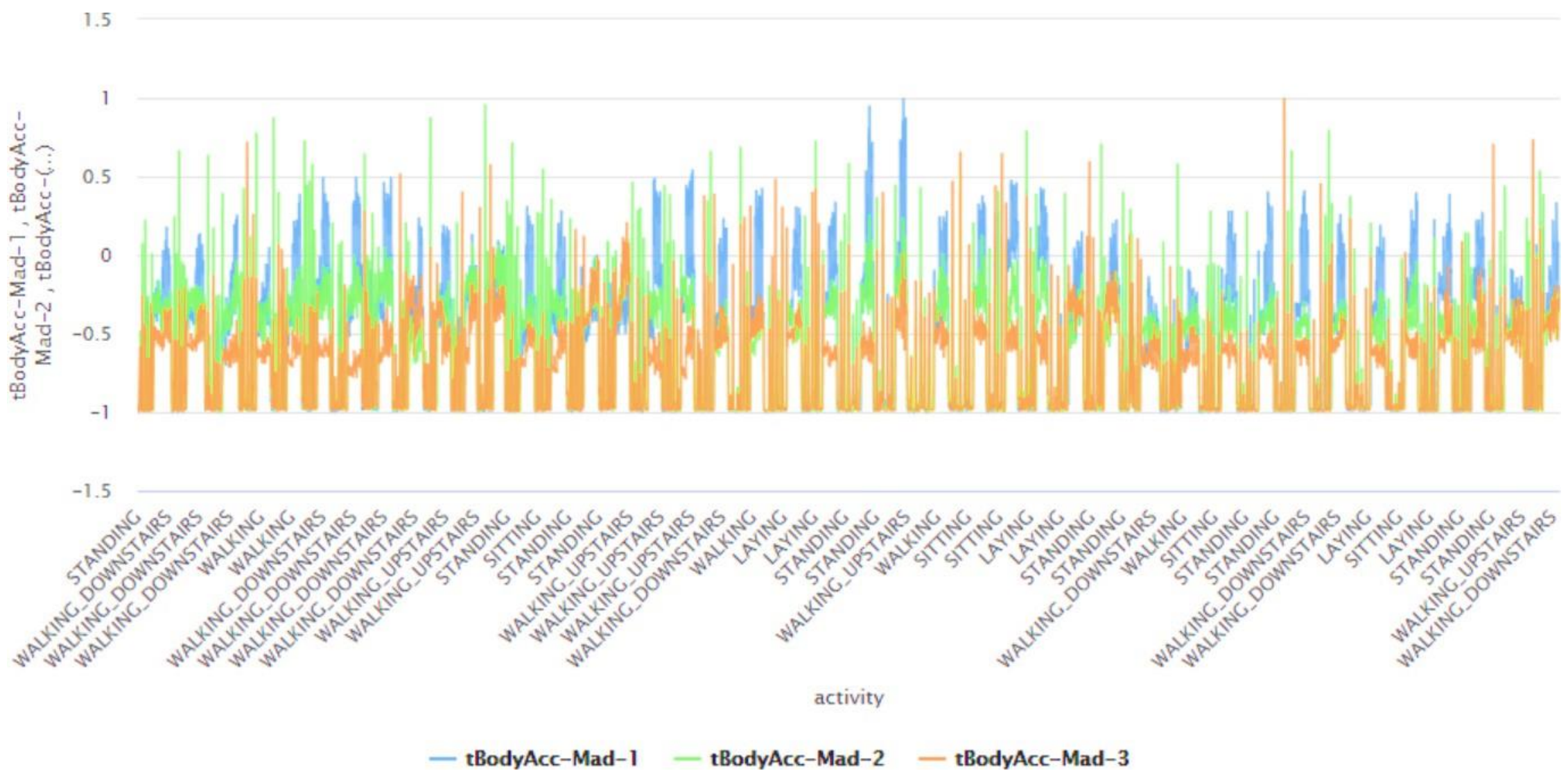

Figure 3: Relationship between tBodyAcc-Mad-1, tBodyAcc-Mad-2, tBodyAcc-Mad-3 with the different activities. For clarity of representation, only a few activity instances from the dataset are shown here.

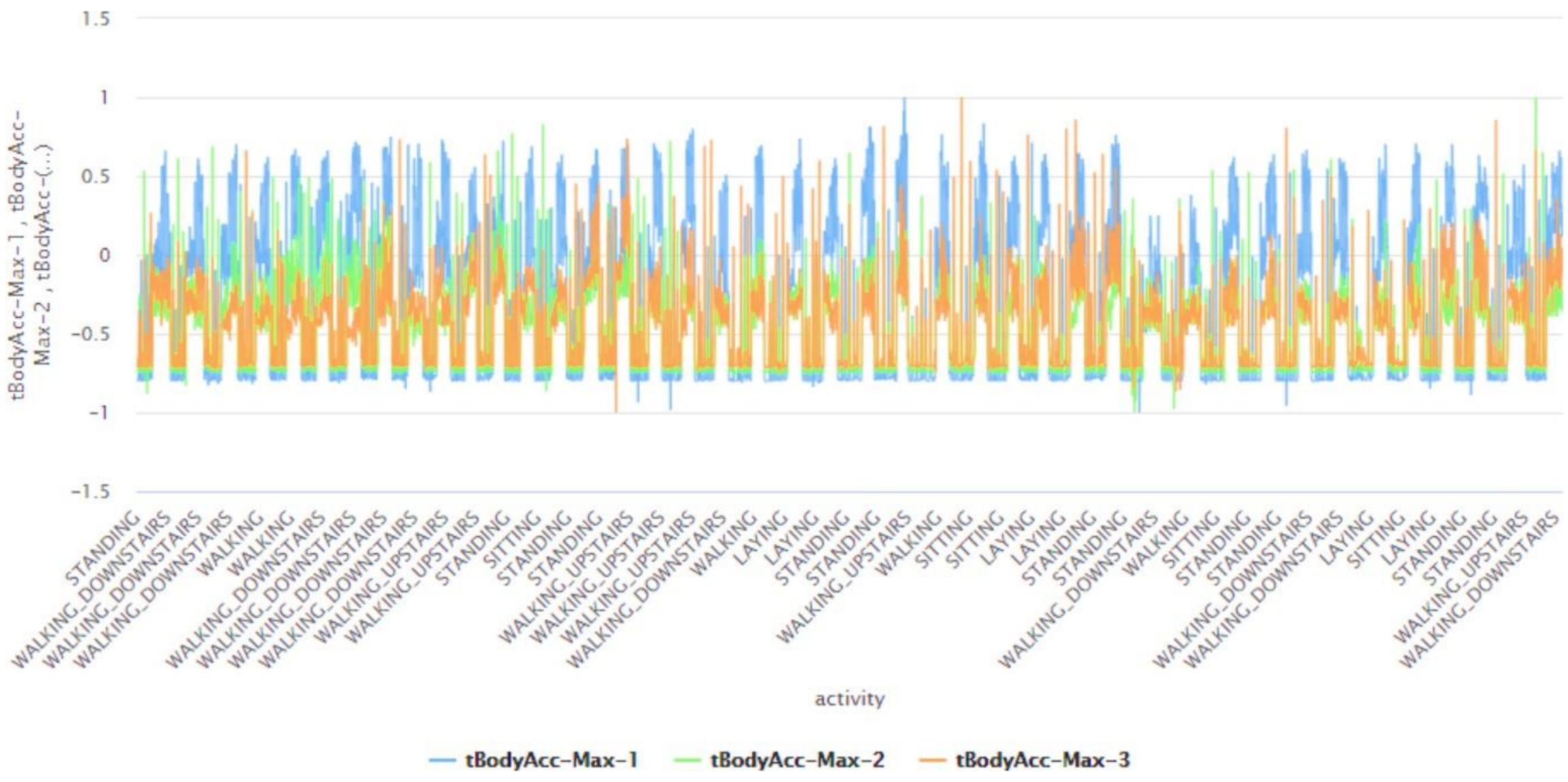

Figure 4: Relationship between tBodyAcc-Max-1, tBodyAcc-Max-2, tBodyAcc-Max-3 with the different activities. For clarity of representation, only a few activity instances from the dataset are shown here.

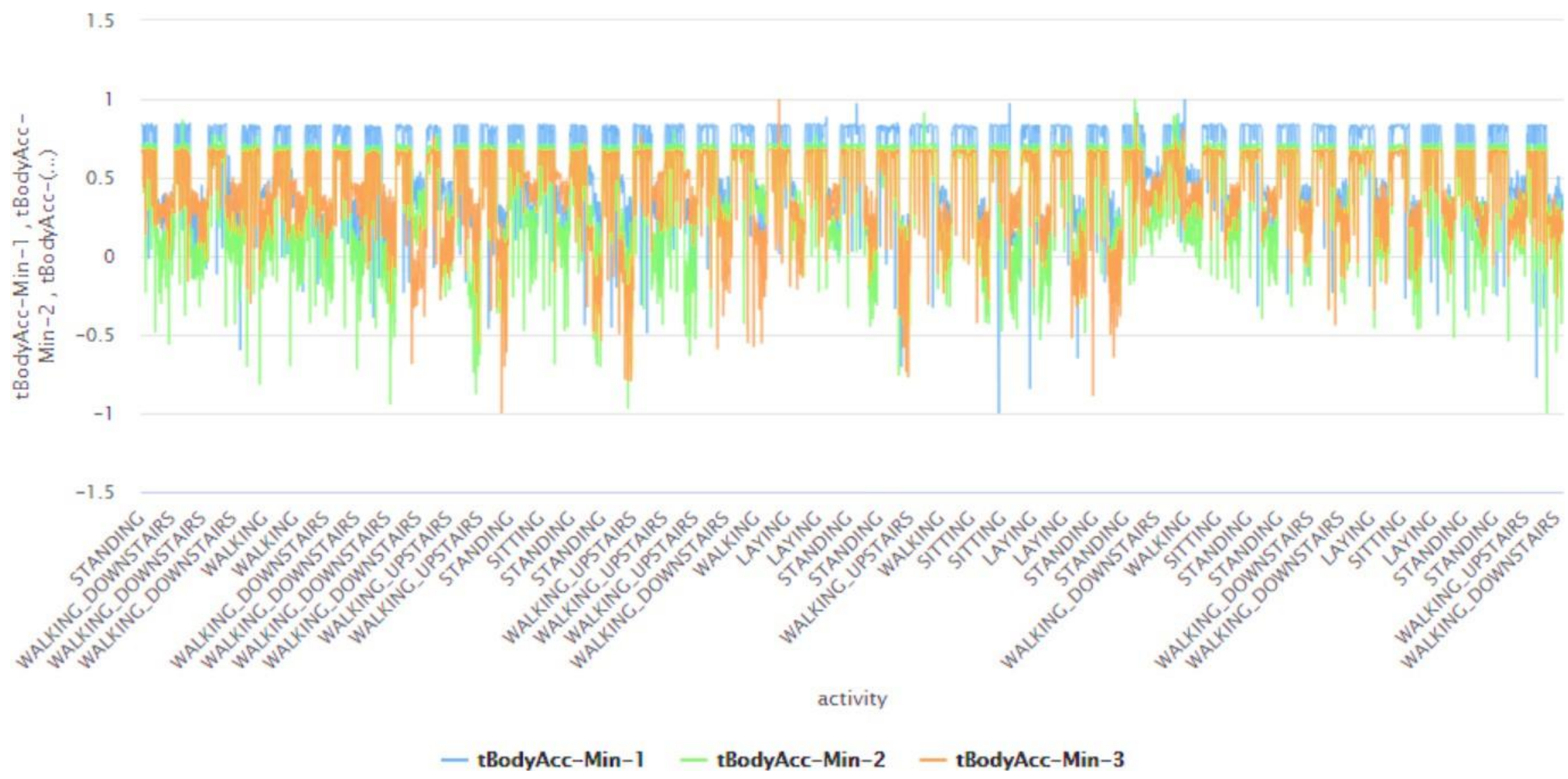

Figure 5: Relationship between tBodyAcc-Min-1, tBodyAcc-Min-2, tBodyAcc-Min-3 with the different activities. For clarity of representation, only a few activity instances from the dataset are shown here.

The value of 'k' for the k-NN classifier was set as 12 as there were 12 labels in the 'activity' attribute. The k-NN 'operator' was customized to use the mixed Euclidean distance approach for the classification. The development of these characteristics of our framework, using RapidMiner, is shown in FIGURES 6-8.

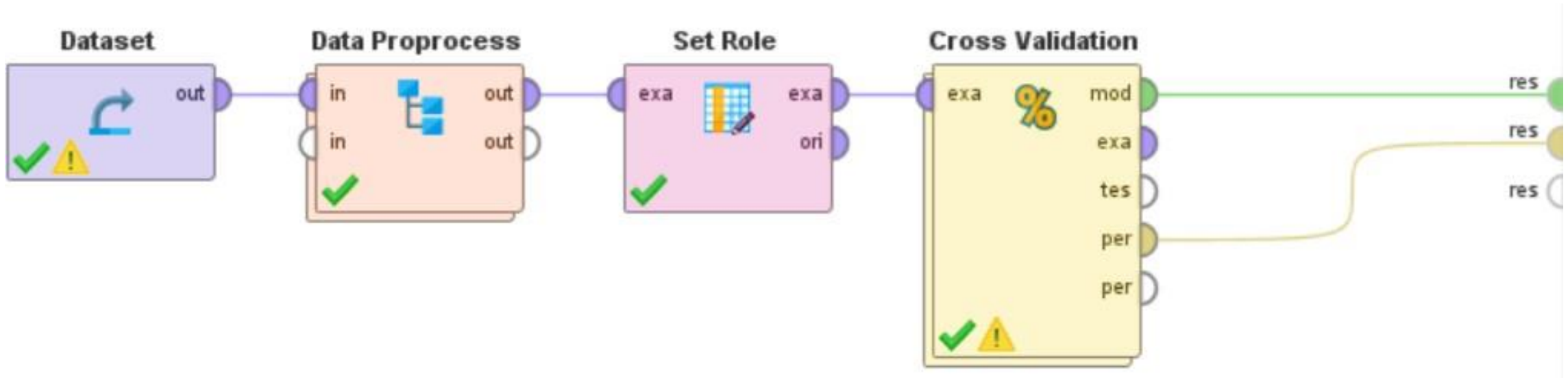

Figure 6: The RapidMiner 'process' of the proposed methodology with the different 'operators' associated with different functionalities. The 'Cross Validation' operator comprised two sub-processes shown in Figures 7 and 8.

The above RapidMiner 'process' for detecting the movements and activities - walking, walking upstairs, walking downstairs, sitting, standing, lying, stand to sit, sit to stand, sit to lie, lie to sit, stand to lie, and lie to stand, achieved an overall accuracy of 82.63%, which is further discussed in Section 5. In Section 5, the findings of the study, which involved developing and comparing the performance characteristics of 17 machine learning approaches to determine the best machine learning approach for this proposed application, are also presented.

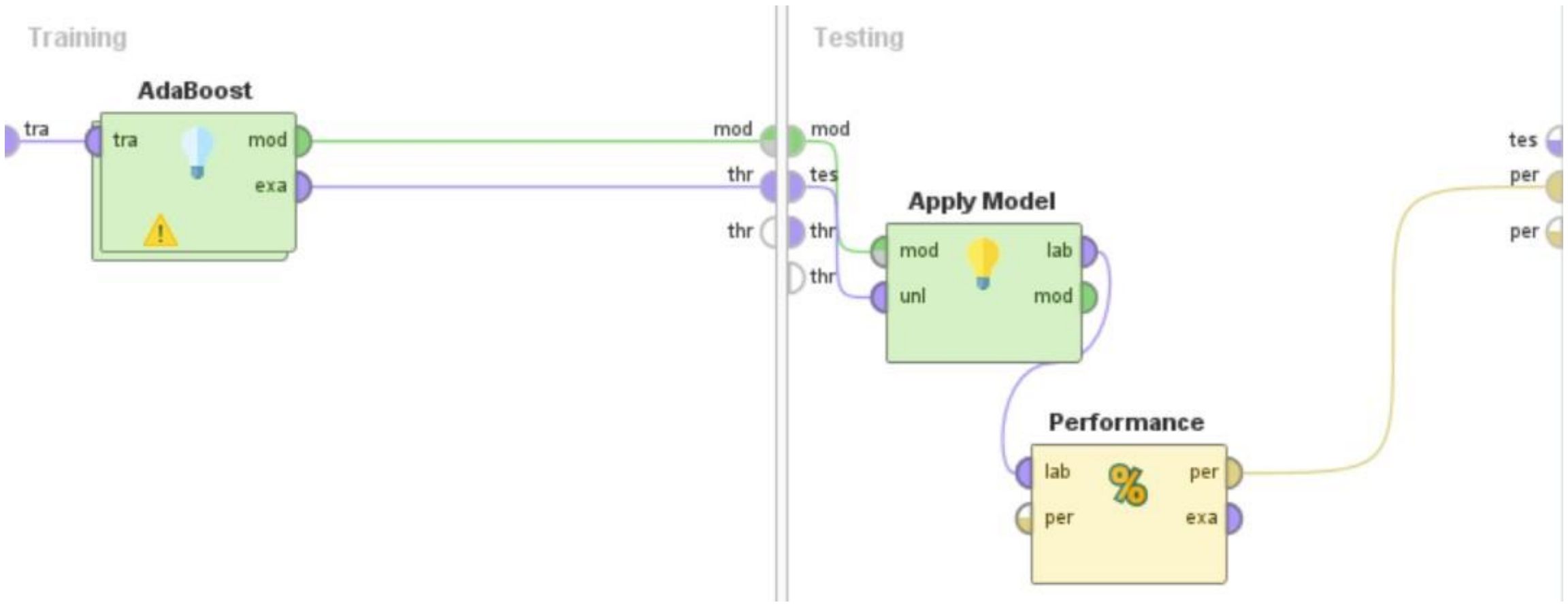

Figure 7: The sub-process of the 'process' shown in Figure 6, that shows the Training and Testing components of the Cross-Validation 'operator', where the AdaBoost was used in the Training component.

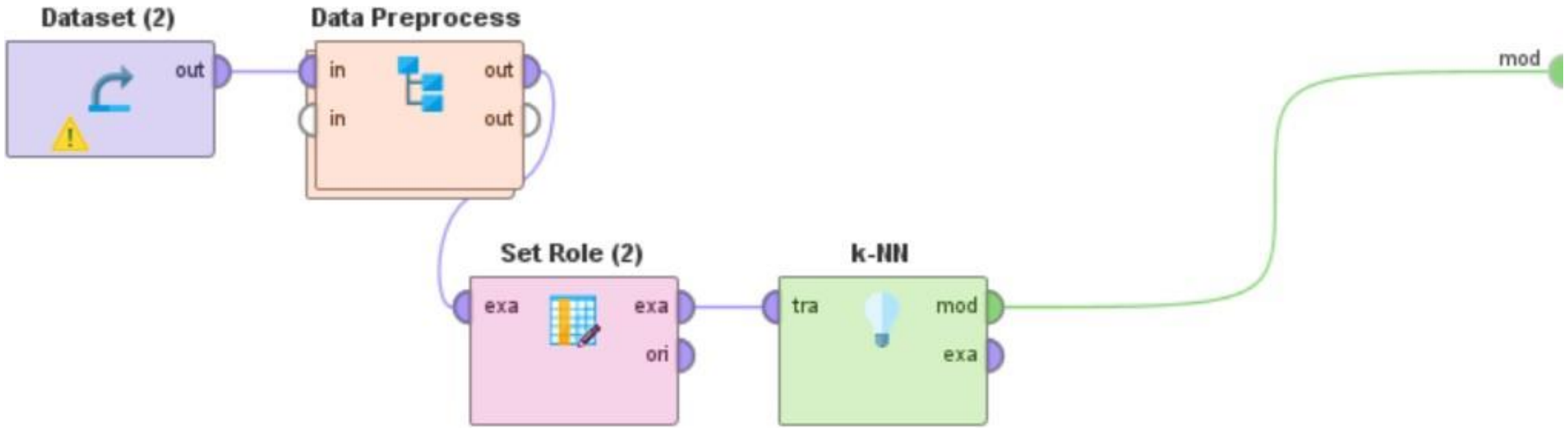

Figure 8: The sub-process of the 'process' shown in Figure 7, where the k-NN-based learning method was developed as a sub-process of the AdaBoost operator.

## 5. RESULTS AND DISCUSSION

We present in this section the results of this research work towards developing a framework for the detection of ADLs and other activities, along with the findings of our comparative study to derive the optimal machine method for the development of this approach. As illustrated in FIGURES 6-8, the proposed system uses a machine learning model consisting of a k-NN classifier with the AdaBoost algorithm, aided by the cross-validation technique. The classifier consisted of 7767 instances with 15 dimensions of the classes walking, walking upstairs, walking downstairs, sitting, standing, lying, stand to sit, sit to stand, sit to lie, lie to sit, stand to lie, and lie to stand. We utilized a confusion matrix to evaluate the performance of our model in detecting these activities and behavioral patterns associated with ADLs.

A confusion matrix is a method of evaluating and studying the performance characteristics of a machine learning-based algorithm [49]. The number of instances of a data label in the predicted class is represented by each row of the matrix, and the number of instances of a data label in the actual class is represented by each column of the matrix. The matrix can also be inverted to have the rows represent the columns and vice versa. Such a matrix allows for the calculation of multiple performance characteristics associated with the machine learning model. These include overall accuracy, individual class precision values, recall, specificity, positive predictive values, negative predictive values, false-positive rates, false-negative rates, and F-1 scores. We utilized two of these characteristics – overall accuracy and class precision, for studying the effectiveness of our proposed framework. The formula to calculate overall accuracy and class precision is shown in Equations (1) and (2).

$$\text{Acc} = \frac{True(P)+True(N)}{True(P)+True(N)+False(P)+False(N)} \quad (1)$$

$$\text{Pr} = \frac{True(P)}{True(P)+False(P)} \quad (2)$$

where Acc is the overall accuracy of the machine-learning model, Pr is the class precision value, True(P) means true positive, True(N) means true negative, False(P) means false positive, and False(N) means false negative. The tabular view and plot view of the confusion matrix are shown in FIGURE 9 and FIGURE 10, respectively.

accuracy: 82.63% +/- 1.52% (micro average: 82.63%)

| | true STANDING | true STAND_TO_SIT | true SITTING | true SIT_TO_STAND | true STAND_TO_LIE | true LAYING | true LIE_TO_SIT | true SIT_TO_LIE | true LIE_TO_STAND | true WALKING | true WALKING_DOWNSTA... | true WALKING_UPSTAIRS | class precision |
|---|---|---|---|---|---|---|---|---|---|---|---|---|---|
| pred. STANDING | 1171 | 1 | 326 | 0 | 0 | 122 | 0 | 0 | 0 | 0 | 0 | 0 | 72.28% |
| pred. STAND_TO_SIT | 0 | 23 | 1 | 0 | 1 | 0 | 0 | 3 | 0 | 0 | 0 | 0 | 82.14% |
| pred. SITTING | 200 | 0 | 848 | 0 | 1 | 138 | 0 | 3 | 0 | 0 | 0 | 0 | 71.26% |
| pred. SIT_TO_STAND | 0 | 0 | 2 | 16 | 0 | 2 | 0 | 0 | 1 | 0 | 0 | 0 | 76.19% |
| pred. STAND_TO_LIE | 0 | 6 | 0 | 0 | 61 | 0 | 0 | 11 | 0 | 0 | 0 | 0 | 78.21% |
| pred. LAYING | 50 | 0 | 114 | 0 | 1 | 1141 | 0 | 0 | 1 | 0 | 0 | 0 | 87.30% |
| pred. LIE_TO_SIT | 0 | 0 | 0 | 0 | 0 | 0 | 46 | 0 | 13 | 0 | 0 | 0 | 77.97% |
| pred. SIT_TO_LIE | 0 | 0 | 0 | 0 | 21 | 4 | 0 | 58 | 0 | 0 | 0 | 0 | 69.88% |
| pred. LIE_TO_STAND | 0 | 0 | 0 | 1 | 0 | 0 | 12 | 0 | 37 | 0 | 0 | 0 | 74.00% |
| pred. WALKING | 2 | 10 | 2 | 5 | 2 | 3 | 0 | 0 | 1 | 1182 | 30 | 112 | 87.62% |
| pred. WALKING_DOWNST... | 0 | 0 | 0 | 0 | 0 | 0 | 0 | 0 | 0 | 8 | 901 | 27 | 96.26% |
| pred. WALKING_UPSTAIRS | 0 | 7 | 0 | 1 | 3 | 3 | 2 | 0 | 4 | 36 | 56 | 934 | 89.29% |
| class recall | 82.29% | 48.94% | 65.58% | 69.57% | 67.78% | 80.75% | 76.67% | 77.33% | 64.91% | 96.41% | 91.29% | 87.05% | |

Figure 9: The tabular view of the confusion matrix showing the performance characteristics of our framework for detecting each activity and behaviors associated with ADLs.

As can be seen from FIGURES 9 and 10, the overall performance accuracy of this learning model was found to be 82.63%. The class precision values for the activity labels - walking, walking upstairs, walking downstairs, sitting, standing, lying, stand to sit, sit to stand, sit to lie, lie to sit, stand to lie, and lie to stand were found to be 72.28%, 82.14%, 71.26%, 76.19%, 78.21%, 87.30%, 77.97%, 69.88%, 74.00%, 87.62%, 96.26%, and 89.29%, respectively. The associated

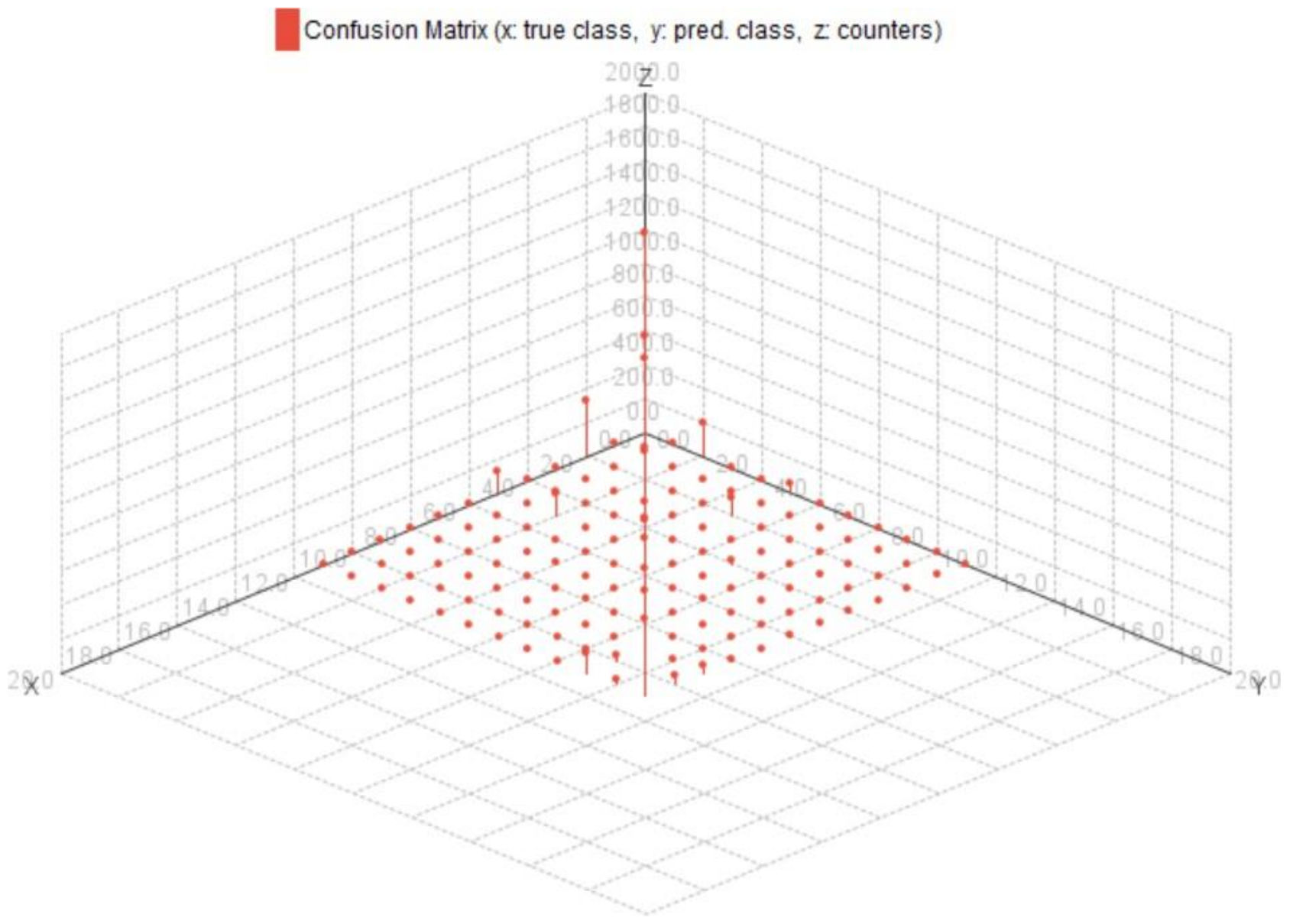

Figure 10: The plot view of the confusion matrix showing the performance characteristics of our framework for detecting each activity and behaviors associated with ADLs.

class recall values for these activity labels were observed as 82.29%, 48.94%, 65.58%, 69.57%, 67.78%, 80.75%, 76.67%, 77.33%, 64.91%, 96.41%, 91.29%, and 87.05%, respectively.

Thereafter, we conducted a comparative case study of 17 machine learning approaches by developing and implementing each of these approaches in our framework to study and deduce the best machine learning approach for the development of this system that can recognize multiple activities and movements associated with ADLs. To perform this case study, we used the same ‘processes’ as shown in FIGURES 5, 6, and 7, the only difference being we replaced the k-NN ‘operator’ in FIGURE 8 with a different ‘operator’ each time that represented a unique learning model from the list of learning models that we included in this case study. This list of these machine learning approaches used in this study, along with the overall accuracy of each approach, is shown in TABLE 1.

If “Per” represents a function that states the overall accuracy of a machine learning model applied on this dataset by following the above-mentioned system architecture, then from TABLE 1, the following may be concluded:

Per(k-NN)>Per(Linear Regression)>Per(Gradient Boosted Trees)>Per(Artificial Neural Network)> Per(AutoMLP)>Per(Linear Discriminant Analysis)>Per(Deep Learning)>Per(Naïve Bayes (Kernel))>Per(Vector Linear Regression)>Per(Naïve Bayes)>Per(Random Forest)>Per(Decision Tree)>Per(Multiway Decision Tree)>Per(Support Vector Machine)>Per(Decision Stump)>Per (Random Tree)>Per(Quadratic Discriminant Analysis).

This comparison clearly shows that the k-NN classifier (used with AdaBoost and k-folds cross-validation) is best suited for the development of this framework, and it outperforms all the other machine learning models. Therefore, the same classification methodology should be used for future work in this field centered around the implementation, deployment, or extension of this framework in real-world settings.

Table 1: Results of the comparative study of different learning approaches (implemented with AdaBoost and Cross-Validation) for detection of multimodal actions and movements associated with ADLs.

| **Learning Approach (with Adaboost and Cross-Validation)** | **Overall Accuracy** |
|---|---|
| Random Forest | 18.32% |
| Artificial Neural Network | 68.11% |
| Decision Tree | 18.32% |
| Multiway Decision Tree | 18.32% |
| Support Vector Machine | 18.32% |
| k-NN | 82.63% |
| Gradient Boosted Trees | 75.02% |
| Decision Stump | 18.32% |
| AutoMLP | 66.62% |
| Linear Regression | 77.80% |
| Vector Linear Regression | 50.66% |
| Random Tree | 18.32% |
| Naïve Bayes | 33.47% |
| Naïve Bayes (Kernel) | 51.71% |
| Linear Discriminant Analysis | 60.78% |
| Quadratic Discriminant Analysis | 18.32% |
| Deep Learning | 55.99% |

## 6. CONCLUSION

As the world's population ages, it creates a variety of needs due to the varying range of bodily limitations associated with the aging process. ADLs (Activities of Daily Living) are essential for one's survival, yet they might be difficult for certain people to execute, especially the elderly people. In the past, experts in this discipline have looked at crutches or wheelchairs as viable answers. However, there are a number of drawbacks to these techniques. Therefore, exoskeleton-based solutions have been investigated by recent researchers; however, there are several research gaps and challenges that still exist. To address these challenges and to contribute towards the independent and assisted living of the elderly, the solutions should be comprehensive, treating bodily movements in the context of tasks performed in ADLs and in terms of body ergonomics, e.g., the extent of the reach of the limbs, shift of weight and center of gravity, and reason for falling out of balance to induce fall. Once all these are taken into account, the key data collected from sensors embedded in wearables and exoskeleton-based solutions to assist and enhance body motions can be studied and analyzed, using Artificial Intelligence and Machine Learning methodologies, which can then used in controlling these solutions; thus, making them smart and adaptable to individuals as per their varying needs. To address these challenges, this paper makes two major scientific contributions to this field.

First, it proposes a machine learning and pattern recognition-based boosted framework that can detect multiple movements and activities associated with ADLs with an accuracy of 82.63%. These activities include walking, walking upstairs, walking downstairs, sitting, standing, lying, stand to sit, sit to stand, sit to lie, lie to sit, stand to lie, and lie to stand. We refer to this approach as 'boosted' as it uses the AdaBoost algorithm with the cross-validation method to train and test the classifier.

Second, it presents and discusses the results of a comprehensive comparative study where we developed and applied 17 different machine learning approaches (with the AdaBoost algorithm and k-folds cross-validation approach) in our proposed framework to determine the best-suited machine learning, in terms of performance characteristics, for development and testing the functionalities of our framework. This study showed that the k-NN classifier is the best-suited machine learning approach as it outperformed all the other 16 machine learning approaches that were developed and studied.

These findings are of paramount importance to the field of exoskeleton research as they would be helpful for making the future of assistive exoskeletons activity and movement-aware so that these systems can be more adaptive and act in an activity-centered and movement-centered manner, thereby increasing the reliance, trust, and acceptance of exoskeletons by elderly and for multiple potential applications. As a specific use case, we also outline how our framework can be seamlessly integrated into the Hybrid Assistive Limb (HAL) exoskeleton [11], a commercially available lower limb exoskeleton. To add we also discuss upon integration of our methodology into the HAL, how the same would be relevant and important to increase the capabilities of the HAL exoskeleton to enable it to support the varied movement and activity-specific needs of diverse end-users – including elderly, disabled, and handicapped people to help them perform ADLs in an independent manner.

As per the best knowledge of the authors, no similar research work has been done in this field so far. As a continuation of this research project, in the near future, we plan to deploy this framework into the HAL exoskeleton and conduct real-time experiments in real-world IoT-based settings with human subjects as per IRB-approved protocols.

## 7. FUNDING

This research received no external funding.

## 8. ACKNOWLEDGEMENTS

The authors would like to thank Isabella Hall, Department of Electrical Engineering and Computer Science at the University of Cincinnati, for her assistance in improving the presentation and formatting of multiple parts of this paper.

## 9. CONFLICTS OF INTEREST

The authors declare no conflict of interest.